# Natural Language Processing: State of The Art, Current Trends and Challenges

Diksha Khurana[1], Aditya Koli[1], Kiran Khatter[1,2] and Sukhdev Singh[1,2]
[1]Department of Computer Science and Engineering
Manav Rachna International University, Faridabad-121004, India
[2]Accendere Knowledge Management Services Pvt. Ltd., India

## Abstract

Natural language processing (NLP) has recently gained much attention for representing and analysing human language computationally. It has spread its applications in various fields such as machine translation, email spam detection, information extraction, summarization, medical, and question answering etc. The paper distinguishes four phases by discussing different levels of NLP and components of **N**atural **L**anguage **G**eneration (NLG) followed by presenting the history and evolution of NLP, state of the art presenting the various applications of NLP and current trends and challenges.

## 1. Introduction

**N**atural **L**anguage **P**rocessing (NLP) is a tract of Artificial Intelligence and Linguistics, devoted to make computers understand the statements or words written in human languages. Natural language processing came into existence to ease the user's work and to satisfy the wish to communicate with the computer in natural language. Since all the users may not be well-versed in machine specific language, NLP caters those users who do not have enough time to learn new languages or get perfection in it.

A language can be defined as a set of rules or set of symbol. Symbol are combined and used for conveying information or broadcasting the information. Symbols are tyrannized by the Rules. Natural Language Processing basically can be classified into two parts i.e. *Natural Language Understanding* and *Natural Language Generation* which evolves the task to understand and generate the text (Figure 1).

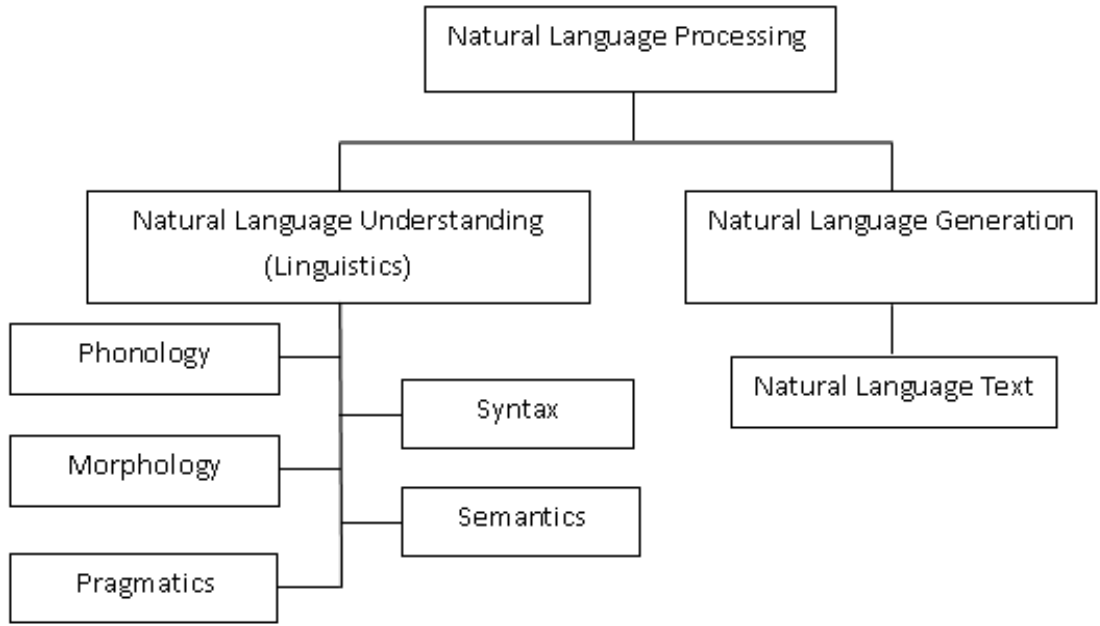

Figure 1. Broad Classification of NLP

**L**inguistics is the science of language which includes *Phonology* that refers to sound, *Morphology* word formation, *Syntax* sentence structure, *Semantics* syntax and *Pragmatics* which refers to understanding.

Noah Chomsky, one of the first linguists of twelfth century that started syntactic theories, marked a unique position in the field of theoretical linguistics because he revolutionised the area of syntax (Chomsky, 1965) [1]. Which can be broadly categorized into two levels Higher Level which include speech recognition and Lower Level which corresponds to natural language. Few of the researched tasks of NLP are Automatic Summarization, Co-Reference Resolution, Discourse Analysis, Machine Translation, Morphological Segmentation, Named Entity Recognition, Optical Character Recognition, Part Of Speech Tagging etc. Some of these tasks have direct real world applications such as Machine translation, Named entity recognition, Optical character recognition etc. *Automatic summarization* produces an understandable summary of a set of text and provides summaries or detailed information of text of a known type. *Co-reference resolution* it refers to a sentence or larger set of text that determines which word refer to the same object. *Discourse analysis* refers to the task of identifying the discourse structure of connected text. *Machine translation* which refers to automatic translation of text from one human language to another. *Morphological segmentation* which refers to separate word into individual morphemes and identify the class of the morphemes. *Named entity recognition* (NER) it describes a stream of text, determine which items in the text relates to proper names. *Optical character recognition* (OCR) it gives an image representing printed text, which help in determining the corresponding or related text. *Part of speech tagging* it describes a sentence, determines the part of speech for each word. Though NLP tasks are obviously very closely interweaved but they are used frequently, for convenience. Some of the task such as automatic summarisation, co-reference analysis etc. act as subtask that are used in solving larger tasks.

The goal of **N**atural **L**anguage **P**rocessing is to accommodate one or more specialities of an algorithm or system. The metric of NLP assess on an algorithmic system allows for the integration of language understanding and language generation. It is even used in multilingual event detection Rospocher et al. [2] purposed a novel modular system for cross-lingual event extraction for English, Dutch and Italian texts by using different pipelines for different languages. The system incorporates a modular set of foremost multilingual Natural Language Processing (NLP) tools. The pipeline integrates modules for basic NLP processing as well as more advanced tasks such as cross-lingual named entity linking, semantic role labelling and time normalization. Thus, the cross-lingual framework allows for the interpretation of events, participants, locations and time, as well as the relations between them. Output of these individual pipelines is intended to be used as input for a system that obtains event centric knowledge graphs. All modules behave like UNIX pipes: they all take standard input, to do some annotation, and produce standard output which in turn is the input for the next module pipelines are built as a data centric architecture so that modules can be adapted and replaced. Furthermore, modular architecture allows for different configurations and for dynamic distribution.

Most of the work in Natural Language Processing is conducted by computer scientists while various other professionals have also shown interest such as linguistics, psychologist and philosophers etc. One of the most ironical aspect of NLP is that it adds up to the knowledge of human language. The field of Natural Language Processing is related with different theories and techniques that deal with the problem of natural language of communicating with the computers. Ambiguity is one of the major problem of natural language which is usually faced in syntactic level which has subtask as lexical and morphology which are concerned with the study of words and word formation. Each of these levels can produce ambiguities that can be solved by the knowledge of the complete sentence. The ambiguity can be solved by various methods such as Minimising Ambiguity, Preserving Ambiguity, Interactive Disambiguity and Weighting Ambiguity [3]. Some of the methods proposed by researchers to remove ambiguity is preserving ambiguity, e.g. (Shemtov 1997; Emele & Dorna 1998; Knight & Langkilde 2000) [3][4][5] Their objectives are closely in line with the last of these: they cover a wide range of ambiguities and there is a statistical element implicit in their approach.

## 2. Levels of NLP

The 'levels of language' are one of the most explanatory method for representing the Natural Language processing which helps to generate the NLP text by realising Content Planning, Sentence Planning and Surface Realization phases (Figure 2).

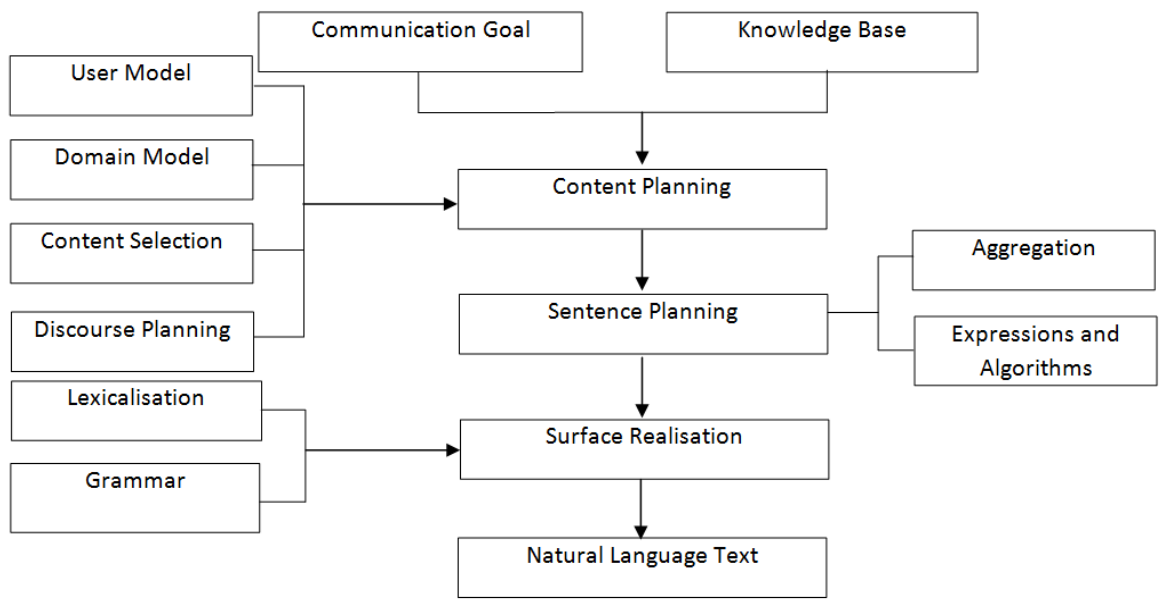

Figure 2. Phases of NLP architecture

Linguistic is the science which involves meaning of language, language context and various forms of the language. The various important terminologies of Natural Language Processing are: -

1. **Phonology**

Phonology is the part of Linguistics which refers to the systematic arrangement of sound. The term phonology comes from Ancient Greek and the term phono- which means voice or sound, and the suffix –logy refers to word or speech. In 1993 Nikolai Trubetzkoy stated that Phonology is "the study of sound pertaining to the system of language". Whereas Lass in 1998 wrote that phonology refers broadly with the sounds of language, concerned with the to lathe sub discipline of linguistics, whereas it could be explained as, "phonology proper is concerned with the function, behaviour and organization of sounds as linguistic items. Phonology include semantic use of sound to encode meaning of any Human language. (Clark *et al.*,2007) [6].

2. **Morphology**

The different parts of the word represent the smallest units of meaning known as Morphemes. Morphology which comprise of Nature of words, are initiated by morphemes. An example of Morpheme could be, the word precancellation can be morphologically scrutinized into three separate morphemes: the prefix pre, the root cancella, and the suffix -tion. The interpretation of morpheme stays same across all the words, just to understand the meaning humans can break any unknown word into morphemes. For example, adding the suffix –ed to a verb, conveys that the action of the verb took place in the past. The words that cannot be divided and have meaning by themselves are called Lexical morpheme (e.g.: table, chair) The words (e.g. -ed, -ing, -est, -ly, -ful) that are combined with the lexical morpheme are known as *Grammatical morphemes* (eg. Worked, Consulting, Smallest, Likely, Use). Those grammatical morphemes that occurs in combination called bound morphemes( eg. -ed, -ing) Grammatical morphemes can be divided into bound morphemes and derivational morphemes.

### 3. Lexical

In Lexical, humans, as well as NLP systems, interpret the meaning of individual words. Sundry types of processing bestow to word-level understanding – the first of these being a part-of-speech tag to each word. In this processing, words that can act as more than one part-of-speech are assigned the most probable part-of speech tag based on the context in which they occur. At the lexical level, Semantic representations can be replaced by the words that have one meaning. In NLP system, the nature of the representation varies according to the semantic theory deployed.

### 4. Syntactic

This level emphasis to scrutinize the words in a sentence so as to uncover the grammatical structure of the sentence. Both grammar and parser are required in this level. The output of this level of processing is representation of the sentence that divulge the structural dependency relationships between the words. There are various grammars that can be impeded, and which in twirl, whack the option of a parser. Not all NLP applications require a full parse of sentences, therefore the abide challenges in parsing of prepositional phrase attachment and conjunction audit no longer impede that plea for which phrasal and clausal dependencies are adequate [7]. Syntax conveys meaning in most languages because order and dependency contribute to connotation. For example, the two sentences: 'The cat chased the mouse.' and 'The mouse chased the cat.' differ only in terms of syntax, yet convey quite different meanings.

### 5. Semantic

In semantic most people think that meaning is determined, however, this is not it is all the levels that bestow to meaning. Semantic processing determines the possible meanings of a sentence by pivoting on the interactions among word-level meanings in the sentence. This level of processing can incorporate the semantic disambiguation of words with multiple senses; in a cognate way to how syntactic disambiguation of words that can errand as multiple parts-of-speech is adroit at the syntactic level. For example, amongst other meanings, 'file' as a noun can mean either a binder for gathering papers, or a tool to form one's fingernails, or a line of individuals in a queue (Elizabeth D. Liddy,2001) [7]. The semantic level scrutinizes words for their dictionary elucidation, but also for the elucidation they derive from the milieu of the sentence. Semantics milieu that most words have more than one elucidation but that we can spot the appropriate one by looking at the rest of the sentence. [8]

### 6. Discourse

While syntax and semantics travail with sentence-length units, the discourse level of NLP travail with units of text longer than a sentence i.e, it does not interpret multi sentence texts as just sequence sentences, apiece of which can be elucidated singly. Rather, discourse focuses on the properties of the text as a whole that convey meaning by making connections between component sentences (Elizabeth D. Liddy,2001) [7]. The two of the most common levels are *Anaphora Resolution* - Anaphora resolution is the replacing of words such as pronouns, which are semantically stranded, with the pertinent entity to which they refer. *Discourse/Text Structure Recognition* - Discourse/text structure recognition sway the functions of sentences in the text, which, in turn, adds to the meaningful representation of the text.

### 7. Pragmatic:

Pragmatic is concerned with the firm use of language in situations and utilizes nub over and above the nub of the text for understanding the goal and to explain how extra meaning is read into texts without literally being encoded in them. This requisite much world knowledge, including the understanding of intentions, plans, and goals. For example, the following two sentences need aspiration of the anaphoric term 'they', but this aspiration requires pragmatic or world knowledge (Elizabeth D. Liddy,2001) [7].

## 3. Natural Language Generation

Natural Language Generation (NLG) is the process of producing phrases, sentences and paragraphs that are meaningful from an internal representation. It is a part of Natural Language Processing and happens in four phases: identifying the goals, planning on how goals maybe achieved by evaluating the situation and available communicative sources and realizing the plans as a text [Figure 3]. It is opposite to Understanding.

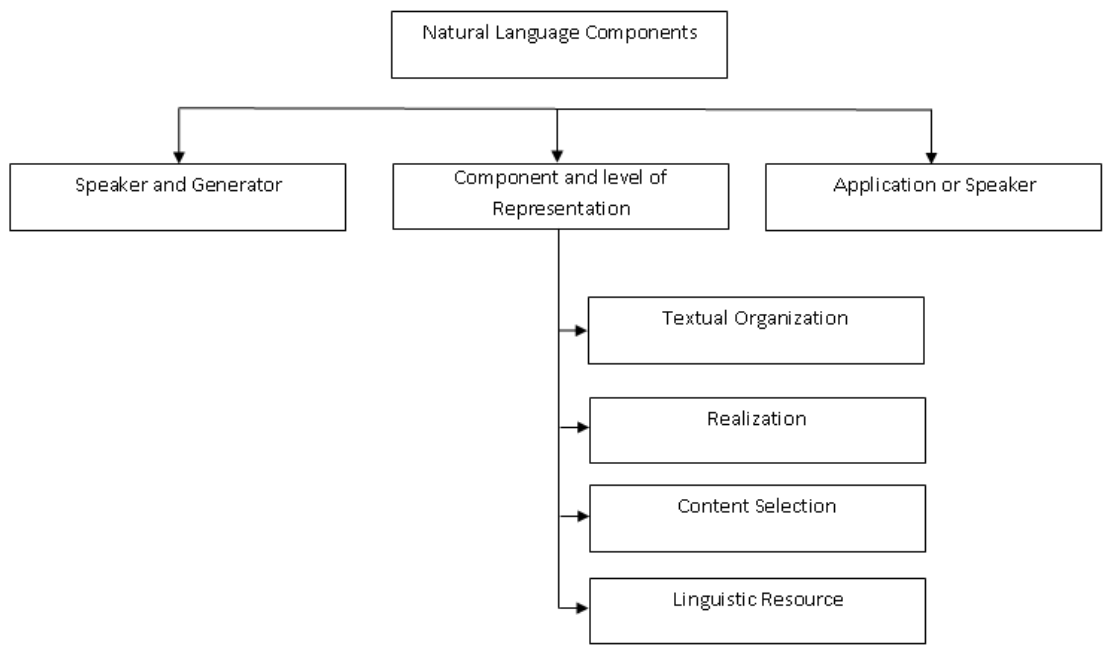

Figure 3. Components of NLG

Components of NLG are as follows:

**Speaker and Generator** – To generate a text we need to have a speaker or an application and a generator or a program that renders the application's intentions into fluent phrase relevant to the situation.

**Components and Levels of Representation** -The process of language generation involves the following interweaved tasks. *Content selection:* Information should be selected and included in the set. Depending on how this information is parsed into representational units, parts of the units may have to be removed while some others may be added by default. *Textual Organization*: The information must be textually organized according the grammar, it must be ordered both sequentially and in terms of linguistic relations like modifications. *Linguistic Resources*: To support the information's realization, linguistic resources must be chosen. In the end these resources will come down to choices of particular words, idioms, syntactic constructs etc. *Realization*: The selected and organized resources must be realized as an actual text or voice output.

**Application or Speaker** – This is only for maintaining the model of the situation. Here the speaker just initiates the process doesn't take part in the language generation. It stores the history, structures the content that is potentially relevant and deploys a representation of what it actually knows. All these form the situation, while selecting subset of propositions that speaker has. The only requirement is the speaker has to make sense of the situation. [9]

## 4. History of NLP

In late 1940s the term wasn't even in existence, but the work regarding machine translation (MT) had started. Research in this period was not completely localised. Russian and English were the dominant languages for MT, but others, like Chinese were used for MT (Booth ,1967) [10]. MT/NLP research was almost died in 1966 according to ALPAC report, which concluded that MT is going nowhere. But later on some MT production systems were providing output to their customers (Hutchins, 1986) [11]. By this time, work on the use of computers for literary and linguistic studies had also started.

As early as 1960 signature work influenced by AI began, with the BASEBALL Q-A systems (Green et al., 1961) [12]. LUNAR (Woods ,1978) [13] and Winograd SHRDLU were natural successors of these systems but they were seen as stepped up sophistication, in terms of their linguistic and their task processing capabilities. There was a widespread belief that progress could only be made on the two sides, one is ARPA Speech Understanding Research (SUR) project (Lea, 1980) and other in some major system developments projects building database front ends. The front-end projects (Hendrix et al., 1978) [14] were intended to go beyond LUNAR in interfacing the large databases.

In early 1980s computational grammar theory became a very active area of research linked with logics for meaning and knowledge's ability to deal with the user's beliefs and intentions and with functions like emphasis and themes.

By the end of the decade the powerful general purpose sentence processors like SRI's Core Language Engine (Alshawi,1992) [15] and Discourse Representation Theory (Kamp and Reyle,1993) [16] offered a means of tackling more extended discourse within the grammatico-logical framework. This period was one of the growing community. Practical resources, grammars, and tools and parsers became available (e.g the Alvey Natural Language Tools (Briscoe et al., 1987) [17]. The (D)ARPA speech recognition and message understanding (information extraction) conferences were not only for the tasks they addressed but for the emphasis on heavy evaluation, starting a trend that became a major feature in 1990s (Young and Chase, 1998; Sundheim and Chinchor ,1993) [18][19]. Work on user modelling (Kobsa and Wahlster , 1989) [20] was one strand in research paper and on discourse structure serving this (Cohen et al., 1990) [21]. At the same time, as McKeown (1985) [22] showed, rhetorical schemas could be used for producing both linguistically coherent and communicatively effective text. Some researches in NLP marked important topics for future like word sense disambiguation (Small et al., 1988) [23] and probabilistic networks, statistically coloured NLP, the work on the lexicon, also pointed in this direction.

Statistical language processing was a major thing in 90s (Manning and Schuetze,1999) [24], because this not only involves data analysts. Information extraction and automatic summarising (Mani and Maybury ,1999) [25] was also a point of focus.

Recent researches are mainly focused on unsupervised and semi-supervised learning algorithms.

## 5. Related Work

Many researchers worked on NLP, building tools and systems which makes NLP what it is today. Tools like Sentiment Analyser, Parts of Speech (POS)Taggers, Chunking, Named Entity Recognitions (NER), Emotion detection, Semantic Role Labelling made NLP a good topic for research.

Sentiment analyser (Jeonghee etal.,2003) [26] works by extracting sentiments about given topic. Sentiment analysis consists of a topic specific feature term extraction, sentiment extraction, and association by relationship analysis. Sentiment Analysis utilizes two linguistic resources for the analysis: the sentiment lexicon and the sentiment pattern database. It analyses the documents for positive and negative words and try to give ratings on scale -5 to +5.

Parts of speech taggers for the languages like European languages, research is being done on making parts of speech taggers for other languages like Arabic, Sanskrit (Namrata Tapswi , Suresh Jain ., 2012) [27], Hindi (Pradipta Ranjan Ray et al., 2003 )[28] etc. It can efficiently tag and classify words as nouns, adjectives, verbs etc. The most procedures for part of speech can work efficiently on European languages, but it won't on Asian languages or middle eastern languages. Sanskrit part of speech tagger is specifically uses treebank technique. Arabic uses Support Vector Machine (SVM) (Mona Diab etal.,2004) [29] approach to automatically tokenize, parts of speech tag and annotate base phrases in Arabic text.

Chunking – it is also known as Shadow Parsing, it works by labelling segments of sentences with syntactic correlated keywords like Noun Phrase and Verb Phrase (NP or VP). Every word has a unique tag often marked as Begin Chunk (B-NP) tag or Inside Chunk (I-NP) tag. Chunking is often evaluated using the CoNLL 2000 shared task.  CoNLL 2000 provides test data for Chunking. Since then, a certain number of systems arised (Sha and Pereira, 2003; McDonald et al., 2005; Sun et al., 2008) [30] [31] [32], all reporting around 94.3% F1 score. These systems use features composed of words, POS tags, and tags.

Usage of Named Entity Recognition in places such as Internet is a problem as people don't use traditional or standard English. This degrades the performance of standard natural language processing tools substantially. By annotating the phrases or tweets and building tools trained on unlabelled, in domain and out domain data (Alan Ritter., 2011) [33]. It improves the performance as compared to standard natural language processing tools.

Emotion Detection (Shashank Sharma, 2016) [34] is similar to sentiment analysis, but it works on social media platforms on mixing of two languages (English + Any other Indian Language). It categorizes statements into six groups based on emotions. During this process, they were able to identify the language of ambiguous words which were common in Hindi and English and tag lexical category or parts of speech in mixed script by identifying the base language of the speaker.

Sematic Role Labelling – SRL works by giving a semantic role to a sentence. For example in the PropBank (Palmer et al., 2005) [35] formalism, one assigns roles to words that are arguments of a verb in the sentence. The precise arguments depend on verb frame and if there exists multiple verbs in a sentence, it might have multiple tags. State-of-the-art SRL systems comprise of several stages: creating a parse tree, identifying which parse tree nodes represent the arguments of a given verb, and finally classifying these nodes to compute the corresponding SRL tags.

Event discovery in social media feeds (Edward Benson et al.,2011) [36], using a graphical model to analyse any social media feeds to determine whether it contains name of a person or name of a venue, place, time etc. The model operates on noisy feeds of data to extract records of events by aggregating multiple information across multiple messages, despite the noise of irrelevant noisy messages and very irregular message language, this model was able to extract records with high accuracy. However, there is some scope for improvement using broader array of features on factors.

## 6. Applications of NLP

Natural Language Processing can be applied into various areas like Machine Translation, Email Spam detection, Information Extraction, Summarization, Question Answering etc.

### 6.1 Machine Translation

As most of the world is online, the task of making data accessible and available to all is a challenge. Major challenge in making data accessible is the language barrier. There are multitude of languages with different sentence structure and grammar. Machine Translation is generally translating phrases from one language to another with the help of a statistical engine like Google Translate. The challenge with machine translation technologies is not directly translating words but keeping the meaning of sentences intact along with grammar and tenses. The statistical machine learning gathers as many data as they can find that seems to be parallel between two languages and they crunch their data to find the likelihood that something in Language A corresponds to something in Language B. As for Google, in September 2016, announced a new machine translation system based on Artificial neural networks and Deep learning . In recent years, various methods have been proposed to automatically evaluate machine translation quality by comparing hypothesis translations with reference translations. Examples of such methods are word error rate, position-independent word error rate (Tillmann et al., 1997) [37], generation string accuracy (Bangalore et al., 2000) [38], multi-reference word error rate (Nießen et al., 2000) [39], BLEU score (Papineni et al., 2002) [40], NIST score (Doddington, 2002) [41]  All these criteria try to approximate human assessment and often achieve an astonishing degree of correlation to human subjective evaluation of fluency and adequacy (Papineni et al., 2001; Doddington, 2002) [42][43].

### 6.2 Text Categorization

Categorization systems inputs a large flow of data like official documents, military casualty reports, market data, newswires etc. and assign them to predefined categories or indices. For example, The Carnegie Group's Construe system (Hayes PJ ,Westein ; 1991)[44] , inputs Reuters articles and saves much time by doing the work that is to be done by staff or human indexers. Some companies have been using categorization systems to categorize trouble tickets or complaint requests and routing to the appropriate desks. Another application of text categorization is email spam filters. Spam filters is becoming important as the first line of defence against the unwanted emails. A false negative and false positive issues of spam filters are at the heart of NLP technology, its brought down to the challenge of extracting meaning from strings of text. A filtering solution that is applied to an email system uses a set of protocols to determine which of the incoming messages are spam and which are not. There are several types of spam filters available. *Content filters*: Review the content within the message to determine whether it is a spam or not. *Header filters*: Review the email header looking for fake information. *General Blacklist filters*: Stopes all emails from blacklisted recipients. *Rules Based Filters*: It uses user-defined criteria. Such as stopping mails from specific person or stopping mail including a specific word. *Permission Filters*: Require anyone sending a message to be pre-approved by the recipient. *Challenge Response Filters*: Requires anyone sending a message to enter a code in order to gain permission to send email.

### 6.3 Spam Filtering

It works using text categorization and in recent times, various machine learning techniques have been applied to text categorization or Anti-Spam Filtering like Rule Learning (Cohen 1996)[45], Naïve Bayes (Sahami et al., 1998 ;Androutsopoulos et al.,2000b ;Rennie .,2000)[46][47][48],Memory based Learning (Androutsopoulos et al.,2000b)[47], Support vector machines (Druker et al., 1999)[49], Decision Trees (Carreras and Marquez , 2001)[50] Maximum Entropy Model (Berger et al. 1996)[51]. Sometimes combining different learners (Sakkis et al., 2001) [52]. Using these approaches is better as classifier is learned from training data rather than making by hand. The naïve bayes is preferred because of its performance despite its simplicity (Lewis, 1998) [53] In Text Categorization two types of models have been used (McCallum and Nigam, 1998) [54]. Both modules assume that a fixed vocabulary is present. But in first model a document is generated by first choosing a subset of vocabulary and then using the selected words any number of times, at least once irrespective of order. This is called Multi-variate Bernoulli model. It takes the information of which words are used in a document irrespective of number of words and order. In second model, a document is generated by choosing a set of word occurrences and arranging them in any order. this model is called multi-nomial model, in addition to the Multi-variate Bernoulli model, it also captures information on how many times a word is used in a document. Most text categorization approaches to anti spam Email filtering have used multi variate Bernoulli model (Androutsopoulos et al.,2000b) [47]

### 6.4 Information Extraction

Information extraction is concerned with identifying phrases of interest of textual data. For many applications, extracting entities such as names, places, events, dates, times and prices is a powerful way of summarize the information relevant to a user's needs. In the case of a domain specific search engine, the automatic identification of important information can increase accuracy and efficiency of a directed search. There is use of hidden Markov models (HMMs) to extract the relevant fields of research papers. These extracted text segments are used to allow searched over specific fields and to provide effective presentation of search results and to match references to papers. For example, noticing the pop up ads on any websites showing the recent items you might have looked on an online store with discounts. In Information Retrieval two types of models have been used (McCallum and Nigam, 1998) [55]. Both modules assume that a fixed vocabulary is present. But in first model a document is generated by first choosing a subset of vocabulary and then using the selected words any number of times, at least once without any order. This is called Multi-variate Bernoulli model. It takes the information of which words are used in a document irrespective of number of words and order. In second model, a document is generated by choosing a set of word occurrences and arranging them in any order. this model is called multi-nomial model, in addition to the Multi-variate Bernoulli model , it also captures information on how many times a word is used in a document

Discovery of knowledge is becoming important areas of research over the recent years. Knowledge discovery research use a variety of techniques in order to extract useful information from source documents like

*Parts of Speech (POS) tagging*, *Chunking or Shadow Parsing*, *Stop-words* (Keywords that are used and must be removed before processing documents), *Stemming* (Mapping words to some base for, it has two methods, dictionary based stemming and Porter style stemming (Porter, 1980) [55]. Former one has higher accuracy but higher cost of implementation while latter has lower implementation cost and is usually insufficient for IR). *Compound or Statistical Phrases* (Compounds and statistical phrases index multi token units instead of single tokens.) *Word Sense Disambiguation* (Word sense disambiguation is the task of understanding the correct sense of a word in context. When used for information retrieval, terms are replaced by their senses in the document vector.)

Its extracted information can be applied on a variety of purpose, for example to prepare a summary, to build databases, identify keywords, classifying text items according to some pre-defined categories etc. For example CONSTRUE, it was developed for Reuters, that is used in classifying news stories (Hayes, 1992) [57]. It has been suggested that many IE systems can successfully extract terms from documents, acquiring relations between the terms is still a difficulty. PROMETHEE is a system that extracts lexico-syntactic patterns relative to a specific conceptual relation (Morin,1999) [58]. IE systems should work at many levels, from word recognition to discourse analysis at the level of the complete document. An application of the Blank Slate Language Processor (BSLP) (Bondale et al., 1999) [59] approach for the analysis of a real life natural language corpus that consists of responses to open-ended questionnaires in the field of advertising.

There's a system called MITA (Metlife's Intelligent Text Analyzer) (Glasgow et al. (1998) [60]) that extracts information from life insurance applications. Ahonen et al. (1998) [61] suggested a mainstream framework for text mining that uses pragmatic and discourse level analyses of text.

### 6.5 Summarization

Overload of information is the real thing in this digital age, and already our reach and access to knowledge and information exceeds our capacity to understand it. This trend is not slowing down, so an ability to summarize the data while keeping the meaning intact is highly required. This is important not just allowing us the ability to recognize the understand the important information for a large set of data, it is used to understand the deeper emotional meanings; For example, a company determine the general sentiment on social media and use it on their latest product offering. This application is useful as a valuable marketing asset.

The types of text summarization depends on the basis of the number of documents and the two important categories are single document summarization and multi document summarization (Zajic et al. 2008 [62]; Fattah and Ren 2009 [63]). Summaries can also be of two types: generic or query-focused (Gong and Liu 2001 [64]; Dunlavy et al. 2007 [65]; Wan 2008 [66]; Ouyang et al. 2011 [67]). Summarization task can be either supervised or unsupervised (Mani and Maybury 1999 [68]; Fattah and Ren 2009 [63]; Riedhammer et al. 2010 [69]). Training data is required in a supervised system for selecting relevant material from the documents. Large amount of annotated data is needed for learning techniques. Few techniques are as follows–

- *Bayesian Sentence based Topic Model (BSTM)* uses both term-sentences and term document associations for summarizing multiple documents. (Wang et al. 2009 [70])
- *Factorization with Given Bases (FGB)* is a language model where sentence bases are the given bases and it utilizes document-term and sentence term matrices. This approach groups and summarizes the documents simultaneously. (Wang et al. 2011) [71])
- *Topic Aspect-Oriented Summarization (TAOS)* is based on topic factors. These topic factors are various features that describe topics such as capital words are used to represent entity. Various topics can have various aspects and various preferences of features are used to represent various aspects. (Fang et al. 2015 [72])

### 6.6 Dialogue System

Perhaps the most desirable application of the future, in the systems envisioned by large providers of end user applications, Dialogue systems, which focuses on a narrowly defined applications (like refrigerator or home theater systems) currently uses the phonetic and lexical levels of language. It is believed that these dialogue systems when utilizing all levels of language processing offer potential for fully automated dialog systems. (Elizabeth D. Liddy, 2001) [7]. Whether on text or via voice. This could lead to produce systems that can enable robots to interact with humans in natural languages. Examples like Google's assistant, Windows Cortana, Apple's Siri and Amazon's Alexa are the software and devices that follow Dialogue systems.

### 6.7 Medicine

NLP is applied in medicine field as well. The Linguistic String Project-Medical Language Processor is one the large scale projects of NLP in the field of medicine [74][75][76][77][78]. The LSP-MLP helps enabling physicians to extract and summarize information of any signs or symptoms, drug dosage and response data with aim of identifying possible side effects of any medicine while highlighting or flagging data items [74]. The National Library of Medicine is developing The Specialist System [79][80][81][82][83]. It is expected to function as Information Extraction tool for Biomedical Knowledge Bases, particularly Medline abstracts. The lexicon was created using MeSH (Medical Subject Headings), Dorland's Illustrated Medical Dictionary and general English Dictionaries. The Centre d'Informatique Hospitaliere of the Hopital Cantonal de Geneve is working on an electronic archiving environment with NLP features [84][85]. In first phase, patient records were archived . At later stage the LSP-MLP has been adapted for French [86][87][88][89] , and finally , a proper NLP system called RECIT [90][91][92][93] has been developed using a method called Proximity Processing [94]. It's task was to implement a robust and multilingual system able to analyze/comprehend medical sentences, and to preserve a knowledge of free text into a language independent knowledge representation [95][96]. The Columbia university of New York has developed an NLP system called MEDLEE (MEDical Language Extraction and Encoding System) that identifies clinical information in narrative reports and transforms the textual information into structured representation [97].

## 7. Approaches

Rationalist approach or symbolic approach assume that crucial part of the knowledge in the human mind is not derived by the sense but is firm in advance, probably by genetic in heritance. Noam Chomsky was the strongest advocate of this approach. It was trusted that machine can be made to function like human brain by giving some fundamental knowledge and reasoning mechanism linguistics knowledge is directly encoded in rule or other forms of representation. This helps automatic process of natural languages. [98] Statistical and machine learning entail evolution of algorithms that allow a program to infer patterns. An iterative process is used to characterize a given algorithm's underlying algorithm that are optimised by a numerical measure that characterize numerical parameters and learning phase. Machine-learning models can be predominantly categorized as either generative or discriminative. Generative methods can generate synthetic data because of which they create rich models of probability distributions. Discriminative methods are more functional and have right estimating posterior probabilities and are based on observations.

Srihari [99] explains the different generative models as one with a resemblance that is used to spot an unknown speaker's language and would bid the deep knowledge of numerous language to perform the match. Whereas discriminative methods rely on a less knowledge-intensive approach and using distinction between language. Whereas generative models, can become troublesome when many features are used and discriminative models allow use of more features. [100] Few of the examples of discriminative methods are Logistic regression and conditional random fields (CRFs), generative methods are Naive Bayes classifiers and hidden Markov models (HMMs).

### 7.1 Hidden Markov Model (HMM)

An HMM is a system where a shifting takes place between several states, generating feasible output symbols with each switch. The sets of viable states and unique symbols may be large, but finite and known. We can descry the outputs, but the system's internals are hidden. Few of the problem could be solved are by Inference A certain sequence of output symbols, compute the probabilities of one or more candidate states with sequences. *Pattern matching* the state-switch sequence is realised are most likely to have generated a particular output-symbol sequence. *Training* the output-symbol chain data, reckon the state-switch/output probabilities that fit this data best.

Hidden Markov Models are extensively used for speech recognition, where the output sequence is matched to the sequence of individual phonemes. Frederick Jelinek, a statistical-NLP advocate who first instigated HMMs at IBM's Speech Recognition Group, reportedly joked, every time a linguist leaves my group, the speech recognizer's performance improves. [101] HMM is not restricted to this application it has several others such as bioinformatics problems, for example, multiple sequence alignment [102]. Sonnhammer mentioned that Pfam hold multiple alignments and hidden Markov model based profiles (HMM-profiles) of entire protein domains. The cue of domain boundaries, family members and alignment is done semi-automatically found on expert knowledge, sequence similarity, other protein family databases and the capability of HMM-profiles to correctly identify and align the members. [103]

### 7.2 Naive Bayes Classifiers

The choice of area is wide ranging covering usual items like word segmentation and translation but also unusual areas like segmentation for infant learning and identifying documents for opinions and facts. In addition, exclusive article was selected for its use of Bayesian methods to aid the research in designing algorithms for their investigation.

## 8. NLP in Talk

This section discusses the recent developments in the NLP projects implemented by various companies and these are as follows:

### 8.1 ACE Powered GDPR Robot Launched by RAVN Systems [104]

RAVN Systems, an leading expert in Artificial Intelligence (AI), Search and Knowledge Management Solutions, announced the launch of a RAVN ("Applied Cognitive Engine") i.e powered software Robot to help and facilitate the GDPR ("General Data Protection Regulation") compliance.

The Robot uses AI techniques to automatically analyse documents and other types of data in any business system which is subject to GDPR rules. It allows users to quickly and easily search, retrieve, flag, classify and report on data mediated to be supersensitive under GDPR. Users also have the ability to identify personal data from documents, view feeds on the latest personal data that requires attention and provide reports on the data suggested to be deleted or secured. RAVN's GDPR Robot is also able to hasten requests for information (Data Subject Access Requests - "DSAR") in a simple and efficient way, removing the need for a physical approach to these requests which tends to be very labour thorough. Peter Wallqvist, CSO at RAVN Systems commented, "GDPR compliance is of universal paramountcy as it will exploit to any organisation that control and process data concerning EU citizens.

**LINK:** http://markets.financialcontent.com/stocks/news/read/33888795/RAVN_Systems_Launch_the_ACE_Powered_GDPR_Robot

### 8.2 Eno A Natural Language Chatbot Launched by Capital One [105]

Capital one announces chatbot for customers called Eno. Eno is a natural language chatbot that people socialize through texting. Capital one claims that Eno is First natural language SMS chatbot from a U.S. bank that allows customer to ask questions using natural language. Customers can interact with Eno asking questions about their savings and others using a text interface. Eno makes such an environment that it feels that a human is interacting. Ken Dodelin, Capital One's vice president of digital product development, said "We kind of launched a chatbot and didn't know it."

This provides a different platform than other brands that launch chatbots like Facebook Messenger and Skype. They believed that Facebook has too much access of private information of a person, which could get them into trouble with privacy laws of U.S. financial institutions work under. Like any Facebook Page admin can access full transcripts of the bot's conversations. If that would be the case then the admins could easily view the personal banking information of customers with is not correct

**LINK:** https://www.macobserver.com/analysis/capital-one-natural-language-chatbot-eno/

## 8.3 Future of BI in Natural Language Processing [106]

Several companies in Bi spaces are trying to get with the trend and trying hard to ensure that data becomes more friendly and easily accessible. But still there is long way for this.BI will also make it easier to access as GUI is not needed. Because now a days the queries are made by text or voice command on smartphones.one of the most common example is Google might tell you today what will be the tomorrows weather. But soon enough, we will be able to ask our personal data chatbot about customer sentiment today, and how do we feel about their brand next week; all while walking down the street. Today, NLP tends to be based on turning natural language into machine language. But with time the technology matures – especially the AI component –the computer will get better at "understanding" the query and start to deliver answers rather than search results.

Initially, the data chatbot will probably ask the question as how have revenues changed over the last three-quarters?' and then return pages of data for you to analyse. But once it learns the semantic relations and inferences of the question, it will be able to automatically perform the filtering and formulation necessary to provide an intelligible answer, rather than simply showing you data.

**Link:** http://www.smartdatacollective.com/eran-levy/489410/here-s-why-natural-language-processing-future-bi

## 8.4 Using Natural Language Processing and Network Analysis to Develop a Conceptual Framework for Medication Therapy Management Research [107]

Natural Language Processing and Network Analysis to Develop a Conceptual Framework for Medication Therapy Management Research describes a theory derivation process that is used to develop conceptual framework for medication therapy management (MTM) research. The MTM service model and chronic care model are selected as parent theories. Review article abstracts target medication therapy management in chronic disease care that were retrieved from Ovid Medline (2000-2016).

Unique concepts in each abstract are extracted using Meta Map and their pairwise cooccurrence are determined. Then the information is used to construct a network graph of concept co-occurrence that is further analysed to identify content for the new conceptual model. 142 abstracts are analysed. Medication adherence is the most studied drug therapy problem and co-occurred with concepts related to patient-centred interventions targeting self-management. The enhanced model consists of 65 concepts clustered into 14 constructs. The framework requires additional refinement and evaluation to determine its relevance and applicability across a broad audience including underserved settings.

**Link:** https://www.ncbi.nlm.nih.gov/pubmed/28269895?dopt=Abstract

### 8.5 Meet the Pilot, world's first language translating earbuds [108]

The world's first smart earpiece Pilot will soon be transcribed over 15 languages. According to Spring wise, Waverly Labs' Pilot can already transliterate five spoken languages, English, French, Italian, Portuguese and Spanish, and seven written affixed languages, German, Hindi, Russian, Japanese, Arabic, Korean and Mandarin Chinese. The Pilot earpiece is connected via Bluetooth to the Pilot speech translation app, which uses speech recognition, machine translation and machine learning and speech synthesis technology.

Simultaneously, the user will hear the translated version of the speech on the second earpiece. Moreover, it is not necessary that conversation would be taking place between two people only the users can join in and discuss as a group. As if now the user may experience a few second lag interpolated the speech and translation, which Waverly Labs pursue to reduce. The Pilot earpiece will be available from September, but can be pre-ordered now for $249. The earpieces can also be used for streaming music, answering voice calls and getting audio notifications.

**Link:** https://www.indiegogo.com/projects/meet-the-pilot-smart-earpiece-language-translator-headphones-travel#/

## REFRENCES